\documentclass[preprint,12pt]{elsarticle}

\usepackage{amssymb}
\usepackage{graphics} 
\usepackage{graphicx} 
\usepackage{color} 
\usepackage[dvipsnames]{xcolor} 
\usepackage{amsmath} 
\usepackage{amssymb}  
\usepackage{bm} 
\usepackage[hidelinks]{hyperref} 
\usepackage{float} 
\usepackage{bookmark} 
\usepackage{algorithm} 
\usepackage{algpseudocode} 
\usepackage{setspace}
\usepackage[latin1]{inputenc} 

\begin{document}

\begin{frontmatter}



\title{Pose Estimation of Vehicles Over Uneven Terrain}
\tnotetext[*]{This work was supported by grant 3-11025 of the Israeli Ministry of Science. 
The generous support of the Paslin Foundation to the Paslin Lab is also acknowledged. }


\author{Yingchong Ma and Zvi Shiller}
\ead{yingchong.ma@hotmail.fr, shiller@ariel.ac.il}

\address{Paslin Laboratory for Robotics and Autonomous Vehicles, Department of Mechanical Engineering and Mechatronics, Ariel University, 40700 Ariel,  Israel.}

\begin{abstract}
This paper presents a method for pose estimation of off-road vehicles moving over uneven terrain. It determines the contact points between the wheels and the terrain, assuming rigid contacts between an arbitrary number of wheels and ground.  The terrain is represented by a 3D points cloud, interpolated by a B-patch to provide a continuous terrain representation.  The pose estimation problem is formulated as a rigid body contact problem for a given location of the vehicle's center of mass over the terrain and a given yaw angle.  The contact points between the wheels and ground are determined by releasing the vehicle from a given point above the terrain, until the contact forces between the wheels and ground, and the gravitational  force, reach equilibrium.  The contact forces are calculated using  singular value decomposition (SVD) of the deduced contact matrix. The proposed method is computationally efficient, allowing real time computation during motion, as demonstrated in several examples. Accurate pose estimations can be used for motion planning, stability analyses and traversability analyses over uneven terrain.

\end{abstract}

\end{frontmatter}



\section{Introduction} \label{sec:intro}

Current work on autonomous vehicles focuses mainly on ideal road conditions \cite{Lav06, ma_2015}. Imperfect roads with bumps and potholes and any other irregular geometry may impose speed limits so as to ensure the vehicle's stability during motion.  The computation of such speed limits is usually treated in the context of off-road motion planning that accounts for vehicle dynamics and surface geometry \cite{shiller75906, mann2006dynamic, Berns2011}. Crucial to this computation is the determination of vehicle pose along the path, particularly when moving at high speeds on uneven terrain, since the rapid changes in the vehicle orientation might cause the vehicle to slide, tip-over, or lose contact with the ground. Furthermore, depending on the terrain geometry, not all wheels may be in contact with the ground at all times, which may severely affect the vehicle's stability and traversability analyses. Developing an efficient algorithm to compute the vehicle pose along a specified path is the focus of this paper.     

Early work on terrain-vehicle interaction focused on developing fundamental models of wheel-terrain interaction \cite{bekker1956theory,bekker1969introduction}. More recently, the issue of computing the vehicle pose on uneven terrain  has been addressed, and efficient numerical approaches were proposed \cite{hal-00557038, howard2007optimal, singh2011planning,  singh2015planning,  jordan2017real, jun2016pose}.  

In \cite{hal-00557038} the vehicle pose is computed by projecting the position of each wheel on the elevation map and assuming that all  wheels contact the ground. Since this method considers only the specfic case where all  wheels are in contact with the ground, it is not applicable to more general cases of multi-wheel vehicles moving over uneven terrain. In \cite{howard2007optimal}, a numerical optimization is used  to minimize the distance between the wheel contact points and the terrain to estimate the vehicle pose. In \cite{singh2011planning} and \cite{singh2015planning} the vehicle is considered as a rigid body with four wheels: three rigid wheels and one compliant wheel. The deflection or extension of the compliant wheel is then computed to determine the vehicle pose. A serious drawback of these methods is that the terrain surface needs to be modeled analytically, which is not always practical.  In \cite{jordan2017real}, a fast pose estimation method is proposed that is based on a digital elevation map.  As these methods consider only 4 contact points, they are not applicable for more general cases.
 
Another approach is proposed in \cite{jun2016pose}, which formulates the problem as a ``linear complementarity problem" (LCP) and  solves it  using the Lemke's method \cite{doi:10.1137/1.9780898719000}. This approach does not require an analytical presentation of the terrain surface and it allows the consideration of multiple contact points. The LCP approach is widely used in contact force computation for rigid bodies \cite{baraff1994fast, lloyd2005fast}. However, these algorithms often fail for large numbers of contact points  \cite{acary2008numerical}, and they require  recomputation of the governing equations after encountering singularities, which slows down computation time \cite{palmerston1984solution}. 

In this paper, the vehicle pose estimation problem over uneven terrain is formulated as a rigid body contact problem between the vehicle body and the terrain. A variety of approaches were developed for computing contact forces. 
Penalty methods are the simplest and earliest \cite{drumwright2008fast}. In these methods, a virtual stiff spring is attached between the contact points and the ground. However this often results in oscillations, and the spring constant is problem specific, which limits the use of these methods in general settings.


We compute the contact forces using Singular Value Decomposition (SVD) of the deduced contact matrix. The advantage of this method is that the procedure of computing the SVD of a matrix is robust \cite{demmel1997applied}, even for a singular matrix, which guarantees that a good approximation always exists even with computation errors and measurement noises. Moreover this approach is fast and simple \cite{press2007numerical2}. 

The paper is structured as follows: Section \ref{sec:terrain} presents the terrain representation and the vehicle model. In Section \ref{sec:robotpose}, the vehicle pose estimation problem is formulated,  and the contact matrix for contact forces computation is deduced. Then, in Section \ref{sec:forces}, contact forces are computed using the SVD based approach. Simulation results are shown thereafter in Section \ref{sec:simulation} to demonstrate the effectiveness of the proposed method.


\section{Terrain Representation and Vehicle Model} \label{sec:terrain}

\subsection{Terrain and Path Representation}
Reliable representation of terrain and traversability is key for most off-road vehicles tasks. In contrast to planar environments in which the terrain and obstacles can be represented as binary data, in 3D cases curvatures and vertical slopes of the uneven road surface must be taken into account for accurate terrain representation.

The majority of the mapping methods project depth information into digital elevation maps by connecting data points 
 by  piecewise continuous cells \cite{Malartre5420701}, which are only  $1^{st}$ order continuous.  A search over such a terrain representation  results in non-smooth paths that do not allow motion at continuous speeds.  

In this paper, the terrain is represented by a cubic B-spline patch  that is interpolated over a  3D point cloud, each point serving as a control point of the B-patch. The B-spline patch representation is smooth and $2^{nd}$ order differentiable, which allows a continuous representation of curvature.  This in turn allows traversing the terrain at continuous velocity profiles.

\begin{figure}[ht!]
\centering
\includegraphics[scale=0.4]{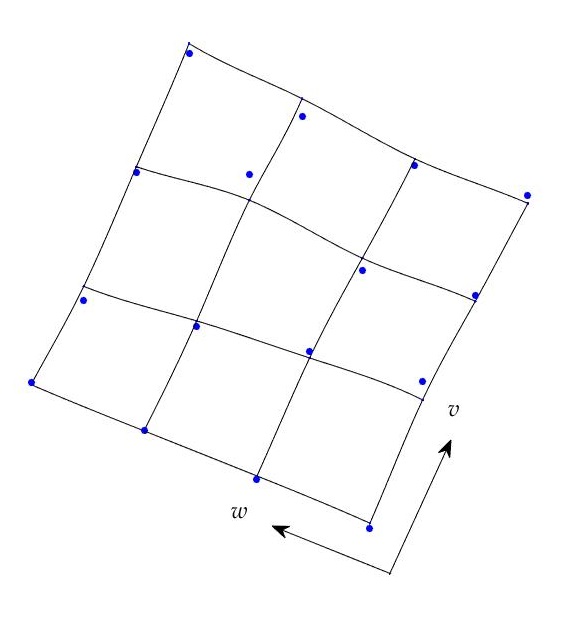}
\caption{A single B-patch terrain with control points}  \label{fig:Bpatch_single}
\end{figure}

Fig. \ref{fig:Bpatch_single} shows a single B-Patch, constructed of 16 control points. A point $\bm{p}$ on the patch  is a function of two parameters $v$ and $w$ \cite{mortensongeometric}:

\begin{equation} \label{eq:bpatch}
\bm{p}(v,w) = \bm{VRD} \bm{R}^{T}\bm{W}^{T},
\end{equation}
where
$$ \bm{V} = [v^3,v^2,v,1],\  v \in [0,1], $$
and
$$ \bm{W} = [w^3,w^2,w,1],\  w \in [0,1],$$
$\bm{R} \in \mathbb{R}^{4 \times 4}$ is a constant matrix specifying the type of spline used to construct the B-spline patch, and $\bm{D} \in \mathbb{R}^{4 \times 4}$ is a matrix of $16$ control points.  Each control point represents an $xy$ location, usually defined as a grid point on a uniform mesh, and a $z$ value that represents its height.  A large surface may be represented by several patches.

\begin{figure}[ht!]
\centering
\includegraphics[scale=0.55]{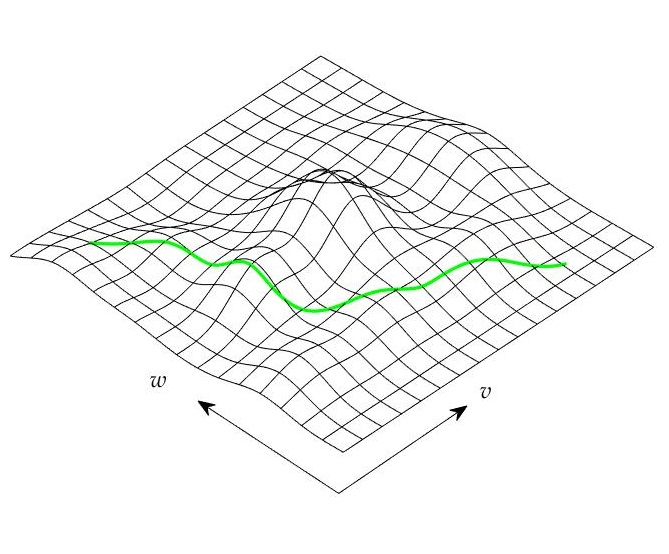}
\caption{A B-spline path over a B-patch terrain}  \label{fig:Bpatch}
\end{figure}

A smooth path over the terrain can be represented by parameterizing $v$ and $w$ with a single parameter $u$:
\begin{equation} \label{eq:path}
\bm{c}(u) = \bm{V}(u)\bm{RDR}^{T}\bm{W}^{T}(u).
\end{equation}
Thus a line in  the $v -w$ space is mapped into a continuous curve on the B-patch. A curve in the $v-w$ space is represented by the B-spline
\begin{equation}
\left[\begin{array}{r}
\bm{V}(u) \\ \bm{W}(u)
\end{array} \right] = \bm{URC}
\end{equation}
where
\begin{equation}
\bm{U} = [u^{3}, u^{2}, u, 1], \ u \in [0, 1]
\end{equation}
$\bm{C} \in \mathbb{R}^{4\times4\times2}$ is an array of control points in the $v-w$ space, and $u$ is the independent parameter along the B spline. B splines are $2^{nd}$ order continuous, which is  essential when computing  time-optimal trajectories. A typical path over the terrain is shown in green in Fig. \ref{fig:Bpatch}.

\subsection{Vehicle Model} \label{sec:model}

\begin{figure}[ht] 
\centering
\includegraphics[scale=0.50]{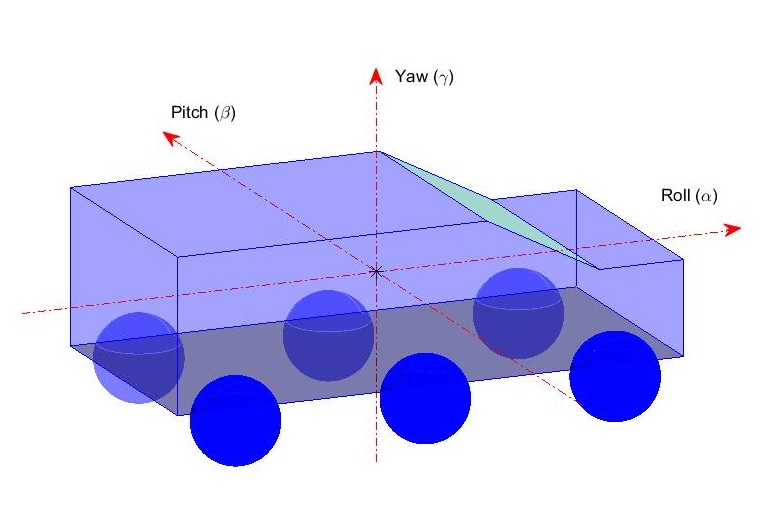}
\caption{A vehicle model}  \label{fig:RobotFrame}
\end{figure}

The  vehicle is assumed to be a rigid body with any number of wheels, each assumed to be rigid, as shown in Fig. \ref{fig:RobotFrame}.  We represent the vehicle configuration by $\bm{r} = (x,y,z,\alpha,\beta,\gamma)$, where $(x,y,z)$ are the coordinate of \textit{the center of mass} of the vehicle body in the inertial frame and $(\alpha,\beta,\gamma)$ are the $roll$, $pitch$ and $yaw$ angles with respect to the vehicle body frame, respectively. 

The number of wheels, their location relative to the vehicle, and the terrain profile, determine the vehicle pose and the contact points, as discussed next.

\section{Pose Estimation} \label{sec:robotpose}

For a given position and orientation $\bm{h}=(x,y,\gamma)$, we wish to find the remaining configuration $\bm{q}=(z,\alpha,\beta)^T$. The vehicle pose is calculated by dropping the vehicle vertically at the configuration $\bm{h}$ under the gravitational force until the vehicle touches the ground, and the gravitational force and the contact forces reach static equilibrium. The wheels in contact are determined from their configuration, location and the ground topography.

Since  $\bm{h}=(x,y,\gamma)$ is fixed, the vehicle motion downward is of three degrees-of-freedom $\bm{q} = (z, \alpha, \beta)^T$. Denoting $\bm{f} \in \mathbb{R}^k$ as the vector of contact forces along the normal direction, $\bm{G} \in \mathbb{R}^k$ as the generalized gravitational force vector, where $k$ is the number of wheels,  the vehicle motion  can be described by:
\begin{equation}   \label{eq:motion}
\bm{M}\ddot{\bm{q}} = \bm{W}_{f}\bm{f} +\bm{W}_{g}\bm{G},
\end{equation} 
where
\begin{equation*}
\begin{array}{ccc}
\bm{M} & = & \left(\begin{array}{ccc}
m & 0 & 0\\
0 & I_{\alpha} & 0\\
0 & 0 & I_{\beta} \\
\end{array} \right),
\end{array} 
\end{equation*}
with $(m,I_{\alpha},I_{\beta})$ being the vehicle mass, the moments of inertia with respect to roll and pitch axes, respectively; $\bm{W}_{f} \in \mathbb{R}^{3 \times k}$ and $\bm{W}_{g} \in \mathbb{R}^{3 \times k}$ are Wrench matrices, which map the normal contact forces and gravitational forces acting on the wheels  to wrenches in the vehicle-body frame.

Equation \eqref{eq:motion} is used to compute the vehicle's trajectory on its way down 
by integrating $\ddot{\bm{q}}$. When there is no contact between the wheels and ground,  $\ddot{\bm{q}}$ can be calculated since the contact forces $\bm{f}$ are zero, and the vehicle is driven only by the gravitational force $\bm{G}$. As soon as one wheel touches the ground, the element in $\bm{f}$ of Eq. \eqref{eq:motion}, corresponding to the contacted wheel, becomes nonzero. We then need to  determine $\bm{f}$ to compute $\ddot{\bm{q}}$.

To solve for $\bm{f}$ at each iteration along the vehicle's vertical trajectory, we first derive an incremental expression for the vehicle's velocity $\bm{v}$, which is then mapped to the rate of change of the minimum vertical distance between each wheel and ground.    

Let's write equation \eqref{eq:motion} in a discrete form:
\begin{equation} \label{eq:discrete}
\bm{M} \bigg( \frac{ \bm{v}^{i} - \bm{v}^{i-1} }{  \Delta t } \bigg) =  \bm{W}_{f}\bm{f}^{i} + \bm{W}_{g}\bm{G},
\end{equation} 
where $\bm{v} = (\dot{z},\dot{\alpha},\dot{\beta})^T$ is the velocity vector, $i$ is the iteration index, and $\Delta t$ is the integration time step. Solving for velocity $\bm{v}^{i}$  yields:
\begin{equation} \label{eq:velocity}
 \bm{v}^{i} =  \bm{v}^{i-1} + \Delta t\bm{M}^{-1} (\bm{W}_{f}\bm{f}^{i} + \bm{W}_{g}\bm{G}).
\end{equation}

We denote $\bm{d}^i$ as the vector of minimum vertical distances between the wheels and the terrain at iteration $i$, $\dot{\bm{d}}^i$ as the rate of change of $\bm{d}^i$. As $\bm{W}_f$ relates between forces at the center of mass and the contact forces at the contacted wheels, we can express $\dot{\bm{d}}^i$ as a function of $\bm{v}^{i}$:
\begin{equation} \label{eq:di1}
\dot{\bm{d}}^{i} = \bm{W}_{f}^{T}\bm{v}^{i}.
\end{equation}
Discretizing $\dot{\bm{d}}^i$ yields:
\begin{equation}\label{eq:di2}
\dot{\bm{d}}^{i} = \frac{\bm{d}^{i} - \bm{d}^{i-1} }{\Delta t}.
\end{equation}
Solving for  $\bm{d}^{i}$ and using \eqref{eq:di1} we get: 
\begin{equation} \label{eq:dis}
\bm{d}^{i} =  \bm{d}^{i-1} + \Delta t \bm{W}_{f}^{T} \bm{v}^{i}.
\end{equation}

Substituting \eqref{eq:velocity} into \eqref{eq:dis}, the vertical distance $\bm{d}^{i}$  is expressed as a function of $\bm{f}^{i}$:
\begin{equation} \label{eq:dis_svd}
\bm{d}^{i} = \bm{A} \bm{f}^{i}+ \bm{b},
\end{equation}
where
\begin{equation*} \label{eq:contact_matrix} 
\bm{A} = \Delta t^2\bm{W}_{f}^{T}\bm{M}^{-1}\bm{W}_{f}
\end{equation*}
and
\begin{equation*}
\bm{b} = \bm{d}^{i-1} + \Delta t\bm{W}_{f}^{T} ( \bm{v}^{i-1} + \Delta t \bm{M}^{-1}\bm{W}_{g}\bm{G} ).
\end{equation*}

Equation \eqref{eq:dis_svd} accounts for the vertical distance $\bm{d}^{i} \in \mathbb{R}^k$ and contact forces $\bm{f}^{i} \in \mathbb{R}^k$ of all $k$ wheels. We wish to solve for  $\bm{f}^{i}$ for a given $\bm{d}^{i}$, then substitute $\bm{f}^i$ into \eqref{eq:motion} to integrate the vehicle's vertical trajectory. 

 Note that if $d_j \neq 0\  (j \in [1,k])$,  then $f_j =0$. 
 As these wheels do not affect the vehicle's motion, we  reduce \eqref{eq:motion} and \eqref{eq:dis_svd} by removing the respective distances and contact forces, thus reducing  the dimensions of $\bm{f}$ and $\bm{d}$ to  $\mathbb{R}^p$.  To not introduce a new notation, we continue using the same notation hereafter so that   $\bm{W}_{f} \in \mathbb{R}^{3 \times p}$, $\bm{d}^{i} \in \mathbb{R}^p$ and $\bm{f}^{i} \in \mathbb{R}^p$, where $p$ is the reduced number of wheels that are in contact with the terrain.

By accounting only for the wheels that are in contact with the ground ($d_j = 0, j \in [1,p]$), and assuming a rigid contact between the wheels and ground, \eqref{eq:dis_svd} reduces to: 
\begin{equation} \label{eq:dis_svd2}
\bm{0} = \bm{A} \bm{f}^{i}+ \bm{b}.
\end{equation}
Solving for $\bm{f}^{i}$ yields:
\begin{equation}\label{eq:force1}
\bm{f}^{i} = \bm{A}^{-1}\bm{b}.
\end{equation}
Due to numerical errors and measurement noise, some $d_j$ are not equal exactly to zero, causing $\bm{A}$ to be near singular and hence not invertible.  
In such cases we set:
\begin{equation}
d_j = 0, \ \text{if} \ d_j<d_{\epsilon},
\end{equation}
where $d_{\epsilon}$ is an arbitrarily small distance threshold. 

Furthermore, since direct inversion of $\bm{A}$ may be very sensitive to numerical errors \cite{gregorcic2001singular}, we  compute its pseudo inverse using a Singular Value Decomposition (SVD)  \cite{ben2003generalized}, which provides a good approximation of $\bm{f}$, as discussed next.


\section{Computation of The Contact Forces} \label{sec:forces}

SVD is a powerful technique for solving sets of ill conditioned equations
or matrices  \cite{demmel1997applied}. Computing SVD of matrices is a simple and robust procedure  even for matrices that are singular.   

For completeness, we briefly describe the computation using the SVD method which is based on the following theorem of linear algebra:
\newtheorem{SVD}{Theorem}
\begin{SVD}
Any matrix $\bm{A} \in \mathbb{R}^{M \times N}$ can be represented by  the following \textit{singular value decomposition} \cite{demmel1997applied}:

\begin{equation} \label{eq:svd}
\bm{A} =\bm{S}\bm{\Sigma}\bm{V}^{T},
\end{equation}
where $\bm{\Sigma}$ is an $N \times N$ diagonal matrix of singular values that are positive or zero:
\begin{equation*}
\bm{\Sigma} =  \left( \begin{array}{cccc}
\sigma_1 &  &  & 0 \\
  & \sigma_2 &  &\\
  &  & ... &  \\
0  &  &  & \sigma_N \\
\end{array} \right),
\end{equation*}
and $\bm{S} \in \mathbb{R}^{M \times N}$ and $\bm{V} \in \mathbb{R}^{N \times N}$ are orthogonal matrices.
\end{SVD}

If $\bm{A}$ is square and non-singular, $\bm{A}^{-1}$ can be easily calculated as \cite{demmel1997applied}:


\begin{equation} \label{eq:svd_inverse}
\bm{A}^{-1} =\bm{V}
\bm{\Sigma}^{-1}
\bm{S}^{T}
\end{equation}
where
\begin{equation*}\label{eq:sigma_inv}
\bm{\Sigma}^{-1} = 
 \left( \begin{array}{cccc}
\frac{1}{\sigma_1} & & & 0 \\
  & \frac{1}{\sigma_2} & &\\
  & & ... & \\
 0  & & & \frac{1}{\sigma_p} \\
\end{array} \right).
\end{equation*}

If any element  $\sigma_j$ is zero, which makes $\bm{\Sigma}$ non-invertible, we set 
\begin{equation} \label{eq:replace}
 \frac{1}{\sigma_{j}} = 0, \  \text{ if } \sigma_j = 0.
\end{equation}
Since $\sigma_j$ may not be exactly equal to zero,  the $\sigma_j$ whose ratios to the largest value $\sigma_{max}$ are smaller than $\epsilon$ are set to zero:
\begin{equation} \label{eq:replace2}
 \frac{1}{\sigma_{j}} = 0, \  \text{ if } \left| \frac{\sigma_j}{\sigma_{max}}\right| < \epsilon.
\end{equation} 
where $\epsilon$ is an arbitrarily small constant, set according to the computer's floating point precision. 

We denote by $\bm{\Sigma}^{\#}$  the matrix resulting after applying \eqref{eq:replace} to  $\bm{\Sigma^{-1}}$,    and  $\bm{A}^{\#}$ as the pseudo inverse of $\bm{A}$ computed by the SVD method.  Eq. \eqref{eq:svd_inverse} then becomes: 
\begin{equation} \label{eq:Apseudo}
\bm{A}^{\#} = \bm{V}\bm{\Sigma}^{\#}\bm{S}^{T}.
\end{equation}
%
%
We can  compute  $\bm{f}^{i}$ \eqref{eq:force1}  using  $\bm{A}^{\#}$:

\begin{equation} \label{eq:force2}
\bm{f}^{i} = -\bm{A}^{\#}\bm{b}.
\end{equation}
Having solved for $\bm{f}^{i}$, we can now compute  $\ddot{\bm{q}}$ from  \eqref{eq:motion} to integrate the vehicle's vertical trajectory until reaching equilibrium at some 
$\bm{q}_e=(z,\alpha,\beta)^T$. 

The procedure to compute the vehicle pose is described in the following Algorithm:

\begin{algorithm}[H]
\renewcommand\baselinestretch{1.3}\selectfont 
\caption{Vehicle Pose Estimation } \label{algorithm}
\begin{algorithmic}[1] 
\State{{\bf Function} $Pose\_Estimation \big( \bm{h}(x,y,\gamma), \bm{p}(v,w) \big)$} 

\State{$i=1$} \Comment{Initialization}

\State{$\ddot{\bm{q}}^0=(-g,0,0)$} \Comment{$g$ is gravitational acceleration}

\While{$\ddot{\bm{q}}_{i-1}\neq 0$} \Comment{$\ddot{\bm{q}}=0$ if vehicle's in equilibrium}

\If{$d_j \neq 0, \forall d_{j} \in \bm{d}^{i-1}$} \Comment{$j \in [1,k]$}

\State{$\bm{f}^{i} = 0$}

\Else

\State{$\bm{A}^{\#} = \bm{S}\bm{\Sigma}^{\#}\bm{V}^{T}$} \Comment{Eq. \eqref{eq:Apseudo}} 

\State{$\bm{f}^{i} = -\bm{A}^{\#}\bm{b}$} \Comment{Eq. \eqref{eq:force2}}

\EndIf

\State{$\ddot{\bm{q}}^{i} =\bm{M}^{-1}( \bm{W}_{f}\bm{f}^{i}+\bm{W}_{g}\bm{G})$} \Comment{Eq. \eqref{eq:motion}}

\State{Compute $\bm{q}^{i}(z,\alpha,\beta)$ using $\ddot{\bm{q}}^{i}$}

\State{Update $\bm{d}^{i}$}

\State{$i = i + 1$}

\EndWhile

\State{Return $\bm{q}^{i-1}(z,\alpha,\beta)$}
\end{algorithmic}
\end{algorithm}


\section{Examples} \label{sec:simulation}

The SVD based algorithm was implemented in MATLAB on an Intel i7-4790 CPU, 3.6GHz desktop computer.  The following examples are demonstrated on the  six-wheel vehicle shown in Fig. \ref{fig:RobotFrame}, with  mass of 500$kg$. The length, width and height  of the vehicle are 1.5$m$, 0.9$m$ and 0.5$m$ respectively. The distance threshold used to identify the contact points  with the terrain is $d_{\epsilon}= 1 cm$. In these examples, the wheels that are in contact with the terrain when the vehicle reaches equilibrium, are marked  red, and those that are not are marked blue. The contact forces are shown as green vectors, emanating from the contact points on the ground.  The  computational efficiency of the SVD based method is later compared to the LCP based method.

\begin{figure}[ht!]
\centering
\includegraphics[scale=0.4]{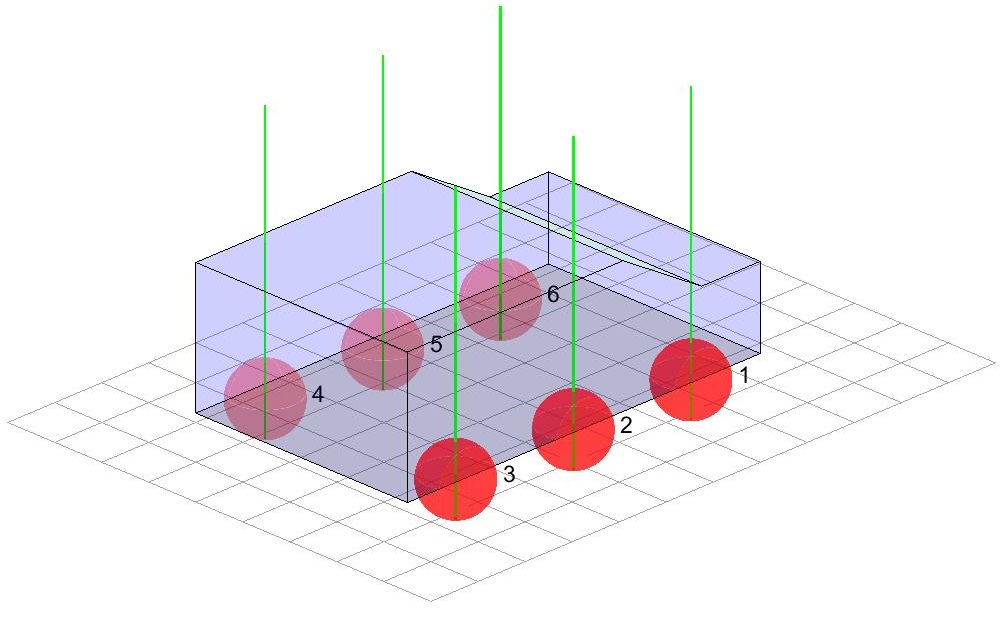}
\caption{Example 1: The six-wheel vehicle is resting on a flat terrain.  the contact forces are shown as vectors emanating from the contact points.} 
\label{fig:SVD_Full}
\end{figure}

\textbf{Example 1}: in this example,  the six-wheel vehicle rests on a flat terrain as shown in Fig. \ref{fig:SVD_Full}.  At equilibrium, the distance between the wheels and ground are set arbitrarily to:
\begin{equation*}
\bm{d}_1 = \left[\begin{array}{rrrrrrrr} 
0.07 & 0.021 &  0.052  & 0.021 & 0.02 &  0.016\\
\end{array}\right] cm.
\end{equation*}
Since all wheels are within the contact distance threshold, their contact forces are nonzero:     
\begin{equation*}
\bm{f}_{S1} = \left[\begin{array}{rrrrrrrr} 
816.7 & 816.7 & 816.7 & 816.7 & 816.7 & 816.7\\
\end{array}\right] N.
\end{equation*}
As expected, the sum of all contact forces is $4900 N$, which  equals  the gravitational force acting on the vehicle.

%

\begin{figure}[ht!]
\centering
\includegraphics[scale=0.4]{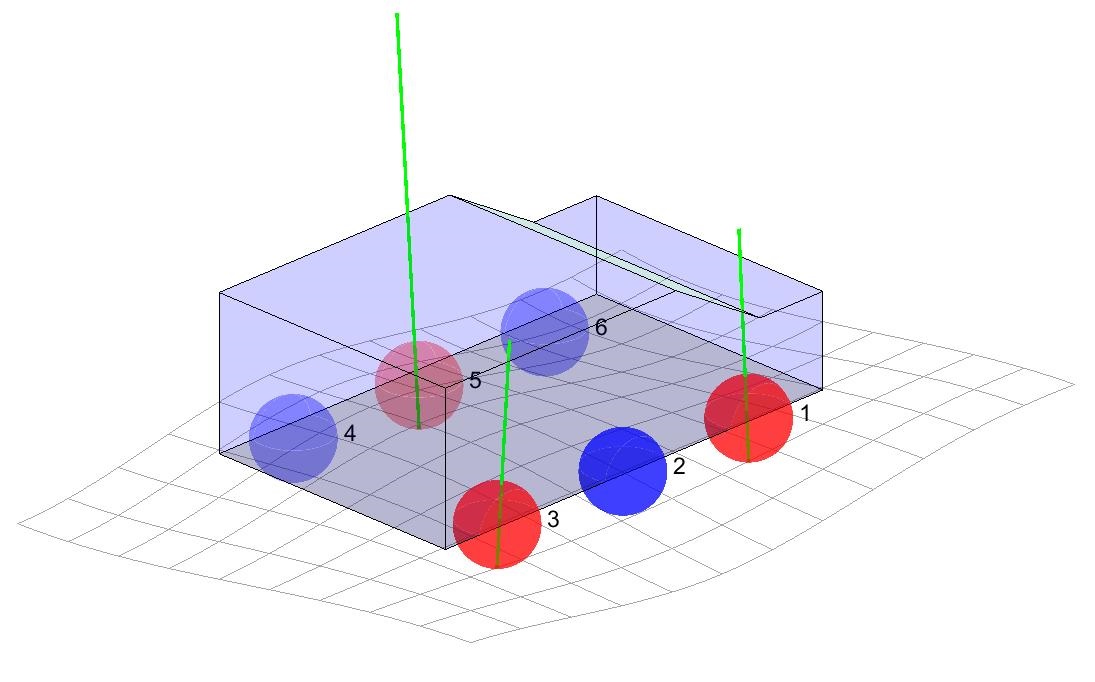}
\caption{Example 2: The six-wheel vehicle is resting on an uneven terrain, with 3 wheels in contact with the ground.} 
\label{fig:SVD_3e}
\end{figure}

\textbf{Example 2}: here,  the six-wheel vehicle rests on an uneven terrain, as shown in Fig. \ref{fig:SVD_3e}.  The distance between the wheels and ground at equilibrium are set  to:
\begin{equation*}
\bm{d}_2 = \left[\begin{array}{rrrrrrrr} 
0.99 & 2.0 &  0.5 & 3 & 0.002 &  2.5\\
\end{array}\right] cm.
\end{equation*}
The  wheels  that 
are within the distance threshold are shown in red   in Fig. \ref{fig:SVD_3e}. Their corresponding contact forces  are:
\begin{equation*}
\bm{f}_{S2} = \left[\begin{array}{rrrrrrrr} 
1225 & 0 & 1225 & 0 & 2450 & 0\\
\end{array}\right] N.
\end{equation*}
Their sum reaches  $4900 N$, as expected.     

 It is interesting to note that although the $1st$ and $3rd$ wheels are within the contact threshold, their contact distance is relatively large compared to the $5th$ wheel, which explains the larger contact force acting on the $5th$ wheel.   

%

\begin{figure}[ht!]
\centering
\includegraphics[scale=0.3]{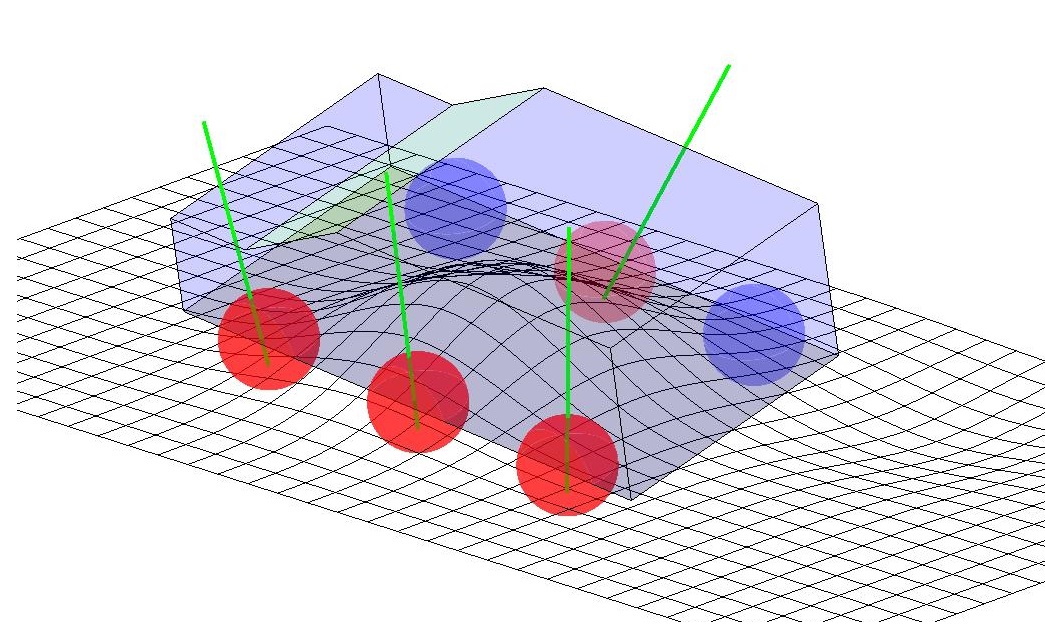}
\caption{Example 3: The six-wheel vehicle over a bump.} 
\label{fig:Rock2_2f}
\end{figure}

\textbf{Example 3}: in this example, the six-wheel vehicle moves over a large bump, as shown in Fig. \ref{fig:Rock2_2f}.  In this case, only four wheels  are in contact with the ground. The computation time from the first contact to equilibrium was $7.6 ms$. An animation of the computation process is shown in \ref{video:bump}.

\begin{figure}[ht!]
\centering
$\begin{array}{cc}
\includegraphics[scale=0.4]{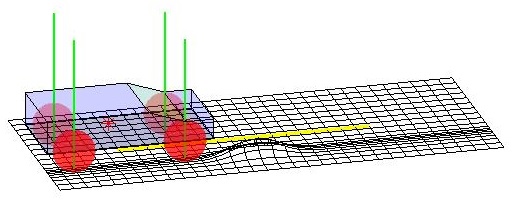} &  \includegraphics[scale=0.37]{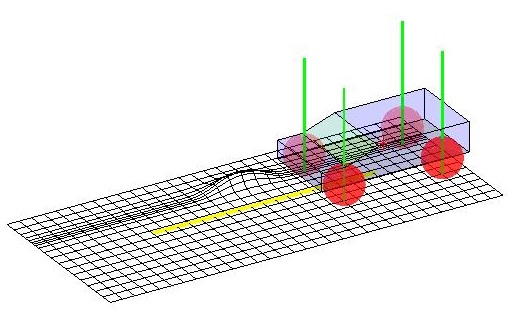} \\
\includegraphics[scale=0.4]{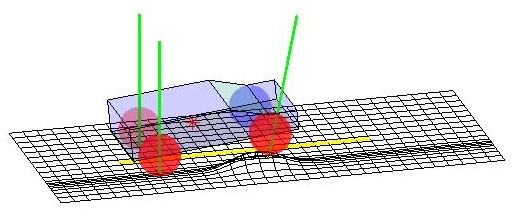} &  \includegraphics[scale=0.37]{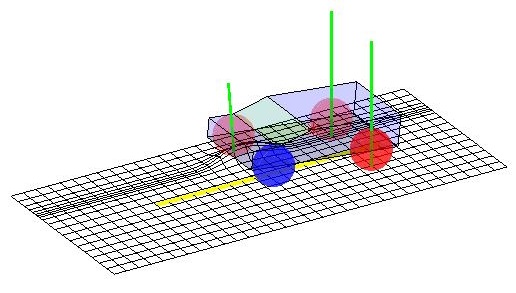} \\
\includegraphics[scale=0.4]{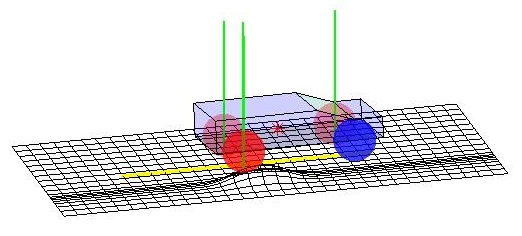} &  \includegraphics[scale=0.37]{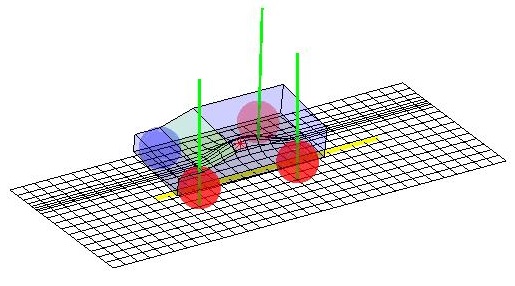} \\
\includegraphics[scale=0.4]{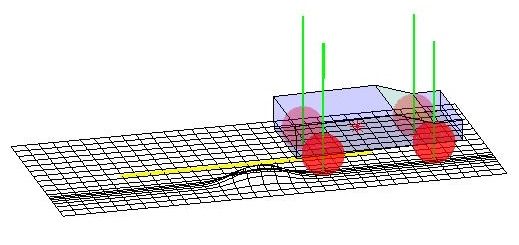} &  \includegraphics[scale=0.37]{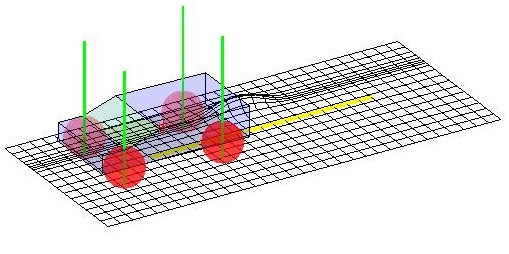} \\
\text{\small Right View} & \text{\small Left View}
\end{array}$
\caption{Example 4: Pose estimation of of a four-wheel vehicle moving over a small rock.} 
\label{fig:Small_Rock}
\end{figure}

\textbf{Example 4}: this example demonstrates  pose estimation along a path segment for a four-wheel vehicle moving over a small rock, as shown in Fig. \ref{fig:Small_Rock}. The contact points switch places as the vehicle is moving over the rock as is shown in Fig. \ref{fig:Small_Rock}, and a video is shown in \ref{video:rock}.

\begin{figure}[ht!]
\centering
\includegraphics[scale=0.3]{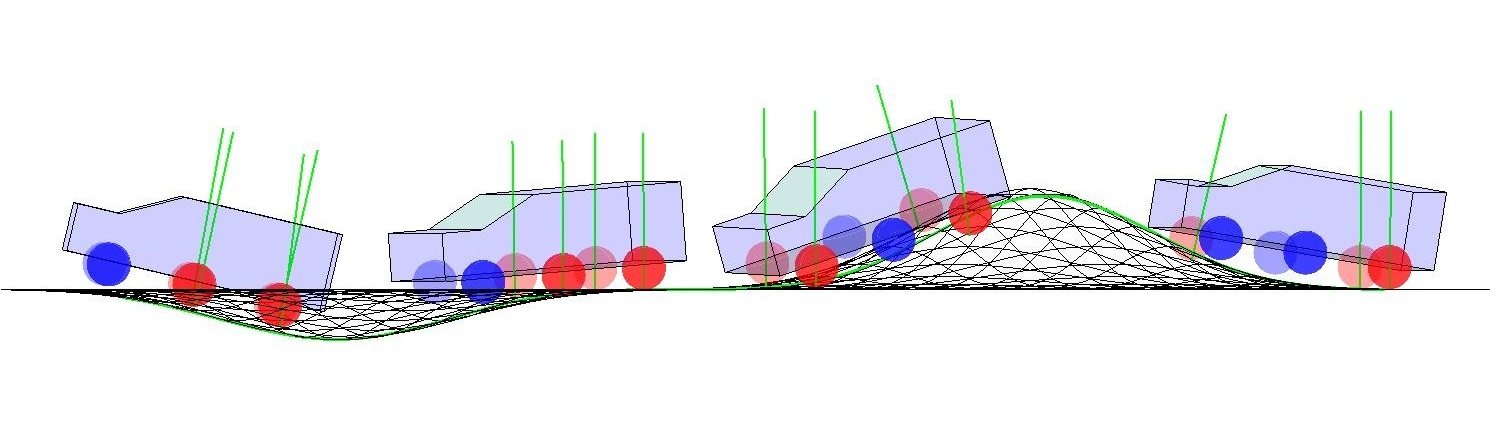}
\includegraphics[scale=0.3]{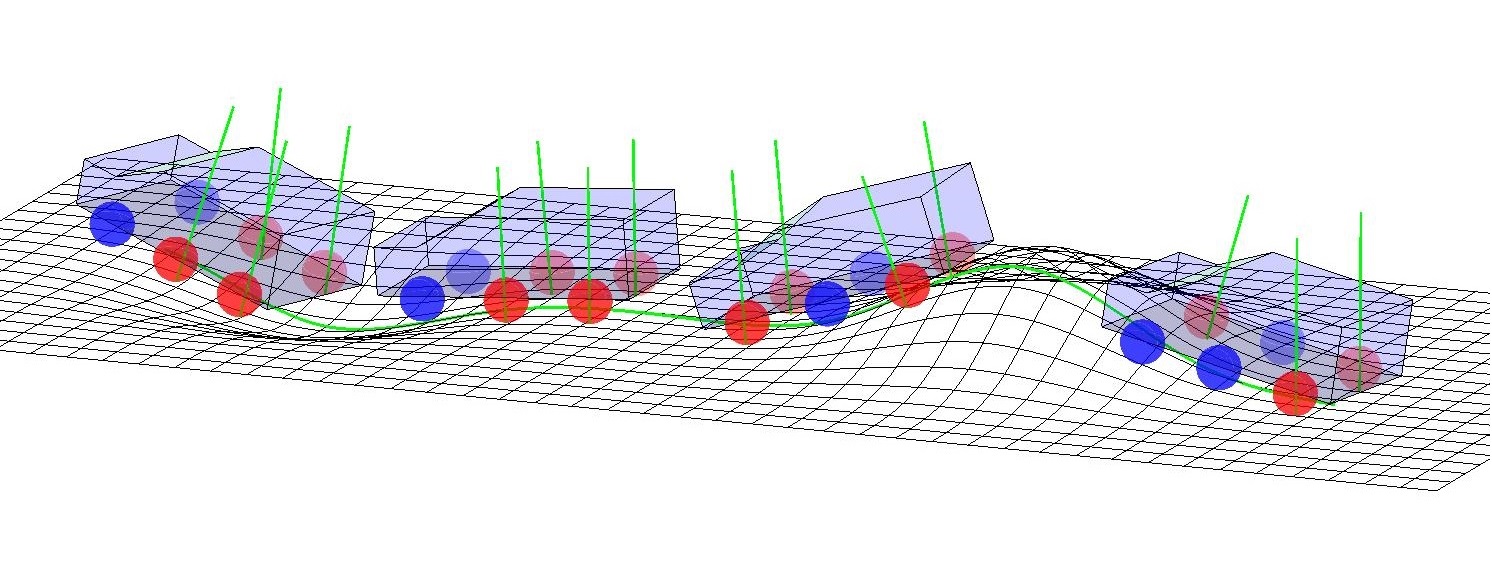}
\caption{Example 5: Pose estimation of a six-wheel vehicle moving over a large bump and a deep hole, shown in two views.} 
\label{fig:Rock1}
\end{figure}

\textbf{Example 5}: this example demonstrates the pose estimation for a six-wheel vehicle moving over a large bump and a deep hole, as demonstrated in Fig. \ref{fig:Rock1}.

\begin{figure}[ht!]
\centering
\includegraphics[scale=0.5]{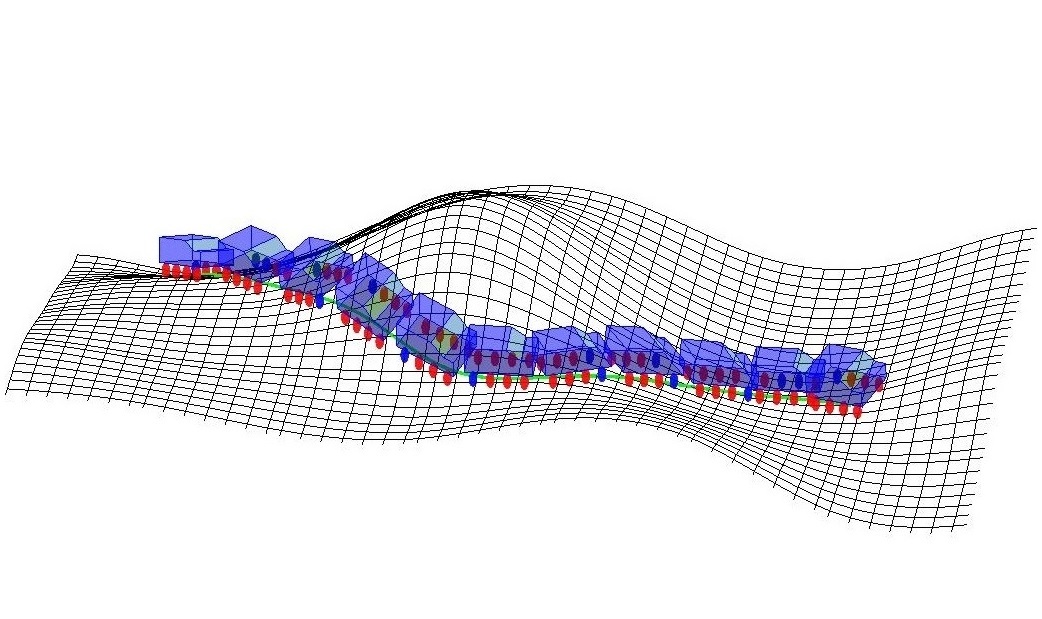}
\caption{Example 6: An eight-wheel vehicle moving along a specified path over uneven terrain.} 
\label{fig:path1CP8}
\end{figure}

\textbf{Example 6}: this example shows an eight-wheel vehicle moving along a path (shown in green) over a mountainous terrain, as shown in Fig. \ref{fig:path1CP8}. The average computation time of one pose from the first contact to equilibrium was $8.1 ms$.

%
%

\subsection{Comparisons to the LCP based method}
To assess the computational efficiency and accuracy of the SVD based approach, we implemented the LCP based method.

The contact forces in example 1, calculated by the LCP based method for the same set  distance $\bm{d}_1$ were
\begin{equation*}
\bm{f}_{L1} = \left[\begin{array}{rrrrrrrr} 
816.4 & 816.4 & 816.4 & 816.4 & 816.4 & 816.4\\
\end{array}\right] N,
\end{equation*}
which is close to the results obtained by the SVD based method. However the computation time of the LCP based method was  $0.9 ms$, compared to  $0.07 ms$ using the SVD method, which is ten times slower.


The contact forces in example 2,  calculated by the LCP based method for the same  distance $\bm{d}_2$ are
\begin{equation*}
\bm{f}_{L2} = \left[\begin{array}{rrrrrrrr} 
1225 & 0 & 1225 & 0 & 2450 & 0
\end{array}\right] N,
\end{equation*}
which is the same as obtained by the SVD method, but took computation time of $1.0 ms$, compared to $0.07 ms$, which is 15 times slower.

It can be seen from these examples that both SVD and LCP based methods provide a good approximation of the contact forces, but the SVD based method is on average 10 times faster than the LCP based method. 

\begin{figure}[h!]
\centering
\includegraphics[scale=0.4]{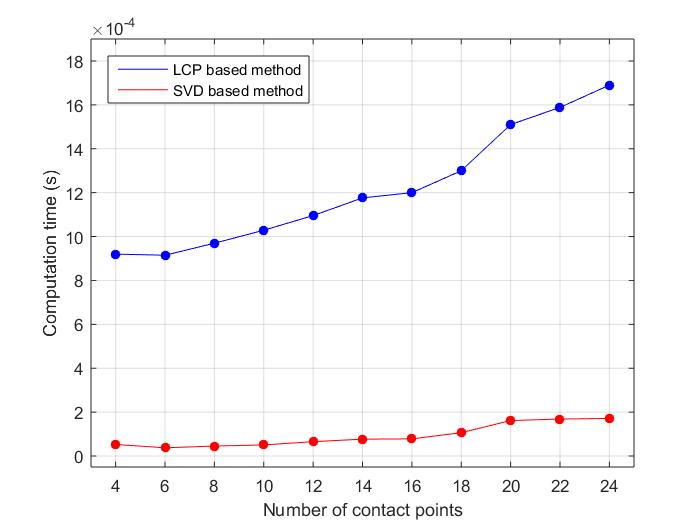}
\caption{Computation time of contact forces (Eq. \eqref{eq:force2}) as a function of the  number of contact points for the SVD and LCP based methods. } 
\label{fig:time}
\end{figure}

Fig. \ref{fig:time} compares  the computation times of both  methods as a function  of the number of wheels. These results were obtained by averaging the computation times of 100 runs for each case for each method. As shown, the computation time of the LCP based method grows from $0.9 ms$ to $1.7 ms$ as the number of wheels increases from $4$ to $24$, whereas the computation time of the SVD approach increased from $0.08 ms$ to $0.18 ms$. It seems that the computation times of both methods are equally affected  by the number of wheels, however, the SVD based approach is consistently more computationally efficient  than the LCP based approach. For large numbers of wheels, the SVD based method is considerably more efficient than the LCP based approach.

\section{Conclusions} \label{sec:conclusion}

A method for pose estimation of off-road vehicles moving over uneven terrain is presented. The terrain is represented by a cubic B-patch that is interpolated over a 3D point cloud. The vehicle pose is calculated by dropping the vehicle vertically at a given configuration under the gravitational force until the contact forces and the gravitational force reach equilibrium. The contact forces between the wheels and the terrain are computed using the Singular Value Decomposition (SVD).

The presented method is robust, easy to implement and computationally efficient, allowing real time computation during motion. The robustness and efficiency of the proposed method is shown through several examples and comparisons. The obtained vehicle poses can be used for motion planning, control design on a rough terrain, stability analyses and traversability analyses over uneven terrain.


\bibliographystyle{elsarticle-num}
\bibliography{PoseEsti}

\appendix
\section{Video for example 3}\label{video:bump} 

[\href{https://youtu.be/-clVfEHUTEQ}{\underline{Video A}}]: Example for a six-wheel vehicle moving over a large bump.

\section{Video for example 4}\label{video:rock}

[\href{https://youtu.be/bjWY9gsjU7M}{\underline{Video B}}]: Example for a four-wheel vehicle moving over a small rock. \label{v:rock}




\end{document}